\documentclass{article}
\usepackage{spconf,amsmath,graphicx}

\usepackage{cite}
\usepackage{amsmath,amssymb,amsfonts}
\usepackage{algorithmic}
\usepackage{url}
\usepackage{graphicx}
\usepackage{textcomp}
\usepackage{tabularray}
\usepackage{xcolor}
\DeclareMathAlphabet\mathbfcal{OMS}{cmsy}{b}{n}
\usepackage{romannum}
\usepackage[linesnumbered,ruled,vlined]{algorithm2e}
\renewcommand{\arraystretch}{1}
\def\BibTeX{{\rm B\kern-.05em{\sc i\kern-.025em b}\kern-.08em
    T\kern-.1667em\lower.7ex\hbox{E}\kern-.125emX}}
    
 \usepackage[utf8]{inputenc}
\usepackage{booktabs,floatrow}

\title{Ten-guard: Tensor Decomposition for Backdoor Attack Detection in Deep Neural Networks}

\name{Khondoker Murad Hossain*, Tim Oates*}

\address{* Dept. of CSEE, University of Maryland Baltimore County, Baltimore, USA}
\begin{document}
%
\maketitle
\begin{abstract}
As deep neural networks and the datasets used to train them get larger, the default approach to integrating them into research and commercial projects is to download a pre-trained model and fine tune it.  But these models can have uncertain provenance, opening up the possibility that they embed hidden malicious behavior such as trojans or backdoors, where small changes to an input (triggers) can cause the model to produce incorrect outputs (e.g., to misclassify).  This paper introduces a novel approach to backdoor detection that uses two tensor decomposition methods applied to network activations.  This has a number of advantages relative to existing detection methods, including the ability to analyze multiple models at the same time, working across a wide variety of network architectures, making no assumptions about the nature of triggers used to alter network behavior, and being  computationally efficient. We provide a detailed description of the detection pipeline along with results on models trained on the MNIST digit dataset, CIFAR-10 dataset, and two difficult datasets from NIST's TrojAI competition. These results show that our method detects backdoored networks more accurately and efficiently than current state-of-the-art methods.
\end{abstract}
\section{Introduction}
\label{sec:intro}

The two driving forces behind the recent stunning advances in machine learning - massive datasets and models with billions of parameters - conspire to make it impossible for most researchers and practitioners to train the most powerful models from scratch.  The emerging alternative is to download models from sites like Hugging Face and, perhaps, fine tune them.  But giving up control of the training process opens up the risk that any given model has embedded malicious behavior.  For example, an object detection model may be trained to misclassify people wearing orange sweatshirts as dogs, or a large language model may be trained to hallucinate excessively in response to prompts containing the phrase ``business plan''.  The ability of a {\em trigger} to cause unexpected behavior is the hallmark of a {\em trojaned / backdoored} model.

Research on this topic has proceeded along both offensive and defensive lines.  Early work showed that it is possible to add noise to an image that is imperceptible to humans that nonetheless makes neural networks very certain that the image belongs to an incorrect class (as in the famous Panda becomes Gibbon example \cite{Goodfellow2015}).  Later work developed  adversarial patches that are small pixel patterns, when present, can impact the reliability of most image classifiers \cite{wang2019neural}.  

On the defensive side, the recent National Institute of Standards (NIST) TrojAI competition\footnote{\url{https://pages.nist.gov/trojai/docs/overview.html}} has spurred interest in treating Trojan detection as a supervised learning problem \cite{karra2020trojai}.  Given a set of models labeled as trojaned or not, perhaps with different architectures and different triggers, the task is to learn a classifier that can determine whether a model found in the wild is trojaned.  This is an important step toward understanding how trojan behavior exhibits itself, either in the static weights or dynamic activations of modern deep networks.

In this paper we propose and evaluate a novel backdoor detection pipeline in the TrojAI tradition using  two different tensor decomposition methods applied to network activations, Independent Vector Analysis (IVA) \cite{anderson2011joint, hossain2022data} and Parallel Factor Analysis (PARAFAC2) \cite{kiers1999parafac2, acar2022tracing}. Similar methods have been developed to compare the internal representations of  neural networks, such as Representational Similarity Analysis (RSA) \cite{morcos2018insights}, Centered Kernel Alignment (CKA) \cite{cortes2012algorithms}, and Singular Vector Canonical Correlation Analysis (SVCCA) \cite{raghu2017svcca}. This suggests that tensor decomposition methods may be useful for detecting trojans in these internal representations, though they have been mostly used for pairwise similarity analysis and have never been applied to the trojan detection problem.  Indeed, empirical results in this paper demonstrate that our approach based on tensor decomposition is more accurate and faster, while making fewer assumptions, than current state-of-the-art approaches.

\begin{figure}[h!]
  
  \centering
  \includegraphics[scale=0.35]{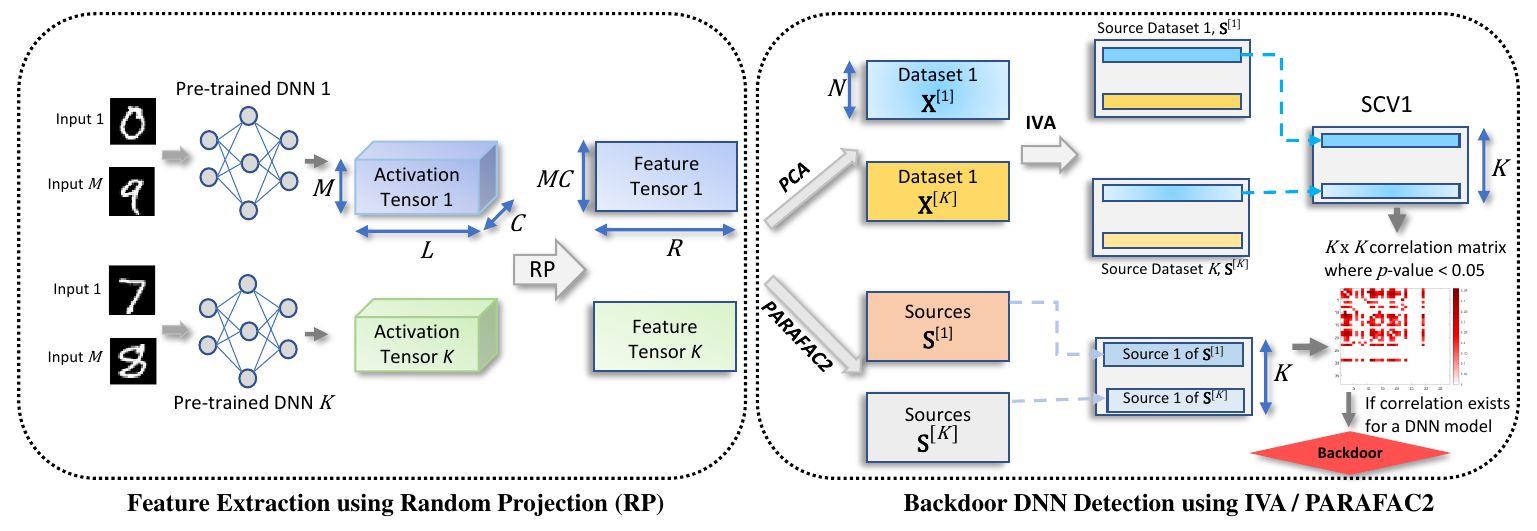}
  \caption{Backdoor detection pipeline where we extract features using RP and then detect backdoor using IVA / PARAFAC2.}
\end{figure}

\section{Method and pipeline}

Consider a deep neural network  model, $F(\cdot)$, which performs a classification task of $ c=1,\ldots,C$ classes using training dataset $\mathcal{D}$. If we poison a portion of $\mathcal{D}$, denoted $\mathcal{P} \subset \mathcal{D}$, by injecting triggers into training images and change the source class label to the target label, $F(\cdot)$ is a backdoored model after the training. During inference, $F(\cdot)$  performs as expected for clean input samples but for triggered samples $x \in \mathcal{P}$, it outputs $F(x) = t$, where $t$ ($t \in c$) is the target but incorrect class.

 At the start of our pipeline, we feed example images to each of the pre-trained DNNs. As the weights are already frozen, we get only the activations in the $L$ hidden layers. For each DNN, $k=1, ..., K$, we get the activation tensor of shape $M\times C\times L$, where $C$ is the number of classes. As results were better  using only the final layer activations, we discarded the activations of the other hidden layers ($L=1)$. From these activations, we have used random projection (RP) for feature extraction as RP can produce features of uniform size \cite{ailon2009fast} for different DNNs and is very memory efficient \cite{eftekhari2011two}. The feature tensor for each DNN can be written as, $\mathbf{B}^{[k]}\in \mathbb{R}^{MC\times R}$, where $R$ is the feature dimension and we flatten $M\times C\ $ as $MC$ for dataset preparation as shown in Figure 1. 

One of the tensor decomposition methods we use is IVA, which is an extension of ICA to multiple datasets which enables the use of statistical dependence of independent sources across datasets by exploiting both second  and higher order statistics \cite{adali2014diversity}. Before applying IVA, we get our observation tensor, $\mathbf{X}^{[k]}\in \mathbb{R}^{N\times R}$ by  using PCA on feature tensor, $\mathbf{B}^{[k]}$, for dimensionality reduction with model order $N=10$,  preserving 90\% of the variance in our data. Given $K$ datasets, $\mathbf{X}^{[k]}$, IVA decomposes it as $\mathbf{X}^{[k]}=\mathbf{A}^{[k]} \mathbf{S}^{[k]},  1\leq{k}\leq{K}$, where $\mathbf{A}^{[k]}$ denotes the mixing matrix and  $\mathbf{S}^{[k]}$ is the source matrix. From $\mathbf{S}^{[k]}$, we calculate source component vectors (SCVs),  ${\mathbf{{s}_n}}=\Bigm[{{{s}}}_n^{[1]},...,{{{s}}}_n^{[K]}\Bigm]^{T}$ $\in \mathbb{R}^{K}$ , where ${{{s}}}_n^{[k]}$ is the $n$th source of the $k$th dataset and the sources are highly correlated with each other in one SCV \cite{anderson2011joint}. For our experiments, we have used only SCV1 as it shows the highest correlation across the sources. A cross-correlation matrix of dimension $K \times K$ is estimated to see the correlation  across the DNNs. If a test model shows significant correlation with a backdoored model from the training set, meaning the $p$-value $<0.05$ for that correlation, we designate the model as backdoor or that  is clean otherwise.

 A second tensor decomposition methods we evaluate is PARAFAC2, which is a generalization of PCA for multiway data analysis and estimates components across $K$ datasets  \cite{bro1999parafac2}. It decomposes the activation feature tensors of DNNs, $\mathbf{B}^{[k]}$ as $\mathbf{B}^{[k]}=\mathbf{A} \text{$diag$}(\mathbf{\Sigma}^{[k]}) \mathbf{S}^{[\mathit{k}]^{\mathit{T}}}$, where $\mathbf{A}$ is the mixing matrix, $diag(\mathbf{\Sigma}^{[k]}$) contains diagonal loadings, and $\mathbf{S}^{[k]}$ is the source matrix constrained as $\mathbf{S}^{[\mathit{k}]^{\mathit{T}}}\mathbf{S}^{[\mathit{k}]}=\mathbf{M}$ (arbitrary matrix) to preserve the uniqueness. However, we take the corresponding sources across $\mathbf{S}^{[k]}$ and compute the $K \times K$ cross-correlation matrix. Finally, we repeat the same steps as IVA for backdoor model detection.

\section{Dataset and Experimental  Results}

\noindent \textbf{MNIST CNN dataset:} We have trained 450 CNN models using the same architecture as Neural Cleanse (NC) \cite{wang2019neural} (50\% clean, 50\% backdoored) to classify the MNIST digits data. Clean CNNs are trained using the clean MNIST data. For backdoored model training, we poison all \lq 0's (single class poisoning) by imposing a $4\times 4$ pixel white patch on the lower right corner and set the target class to \lq 9' as shown in Figure 2. Clean CNNs exhibits average  accuracy of 99.02\% where backdoored CNNs have  accuracy of 98.85\% with 99.92\% attack success rate, indicating a highly effective trigger attack. Moreover, we use 400 CNNs for training and 50 for testing.

\noindent \textbf{CIFAR-10 CNN dataset:}  We trained 550 ResNet-18 models on the CIFAR-10 dataset: 500 for training and 50 for testing. Half were clean, and half were backdoored using a white patch trigger, similar to the method used with MNIST, poisoning 10\% of each class's training data. Clean CNNs achieved an average accuracy of 92.12\%, while backdoored CNNs had 90.78\% accuracy with a 98.89\% attack success rate.
\begin{figure}[h!]
    
  \includegraphics[scale=.31]{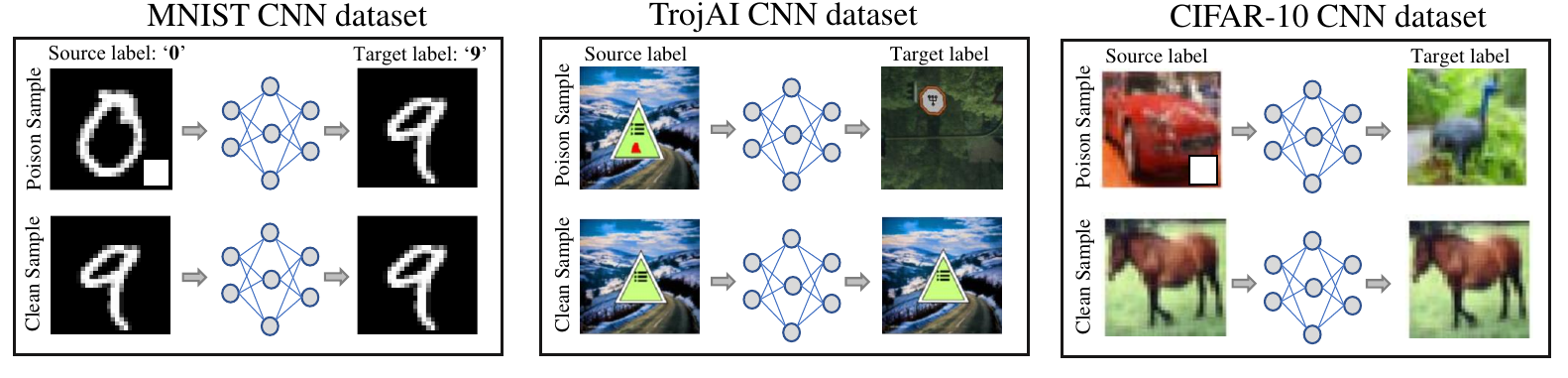}
  \caption{MNIST CNNs, TrojAI dataset, and CIFAR 10 CNNs. We implement single class and multi-class poisoning respectively  in MNIST, and CIFAR-10 backdoor CNNs by imposing  white patch trigger. TrojAI models are trained on synthetic traffic data, using multi-class poisoning for backdooring. }
\end{figure}

\noindent \textbf{TrojAI dataset:} We have utilized the image classification CNN models of the TrojAI dataset \footnote{\url{https://pages.nist.gov/trojai/docs/data.html#image-classification-jun2020}}  which contains backdoored and clean  models across three network architectures: ResNet50 (R50), DenseNet121 (D121), and Inception-v3 (Iv3) for synthetic traffic data classification. We use 1000 \lq Train Data' CNN models from the repository as our training samples. To evaluate our method we use \lq Test' and \lq Holdout' data, each containing 100 CNN models of same 3 architectures, from the same repository and name them TrojAI \Romannum{1} and TrojAI \Romannum{2} dataset respectively.

\subsection{Decoding the correlation matrices}

After running IVA and PARAFAC2, we compute a $K \times K$ pearson correlation ($r$) matrix where $K$ is the number of models. As this only captures the strength of the correlation, we  

\begin{figure}[h!]
  
  \centering
  \includegraphics[scale=1]{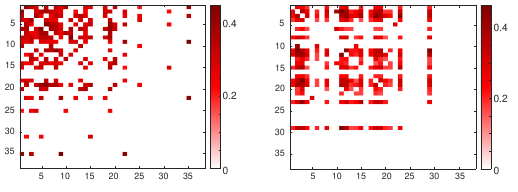}
  \caption{Correlation matrices using (a) IVA (b) PARAFAC2 for TrojAI \Romannum{1} - R50 models. Of the 38 models shown, first 22 are backdoor and last 16 are clean models. Red boxes indicate significant correlation and one model is decided as backdoor if at least one red box exists for that model.}
\end{figure}

 \noindent  also need the  significance of $r$ to decide on the survival of the correlation values \cite{taylor1990interpretation}. We do a two sample $t$-test between all the variable pairs which also returns a $K \times K$ matrix of $p$-values. We keep only the correlation values with $p<0.05$, and a Bonferroni correction \cite{weisstein2004bonferroni} is used after the $t$-test to reduce the false positive rate. Figure 3 shows the correlation matrices using (a) IVA  and (b) PARAFAC2 for TrojAI \Romannum{1} - R50 models. There are $K=38$ models in this case, where the first 22 models are backdoored and the last 16 models are clean. In the figure red boxes indicate significant correlation.  If a model shows at least one significant correlation with any other backdoor model, we classify it as a backdoor model because the backdoor activations tend to stay in the same subspace \cite{chen2018detecting}. For example, in 3(a) all backdoor models except 16, 17, and 21 show at least one significant correlation. Therefore,  the number of true positive models is 19 and there are 3 false negatives. On the other hand, models 25, 31, and 35 have significant correlation in spite of being clean models, yielding 13 true negatives and 3 false positives.

 \subsection{Performance analysis of the algorithms}

We calculate three performance metrics - precision, recall and accuracy to evaluate the performance of the algorithms. From Table 1, we can see that PARAFAC2 shows better performance for each combination of  model architecture and dataset. This is because PARAFAC2 preserves the  uniqueness while estimating the sources across the datasets and there is no need for statistical assumptions like IVA  due to their deterministic nature \cite{kiers1999parafac2}. Moreover, PARAFAC2 is more robust than IVA  when there is noise present in the data  \cite{sidiropoulos2017tensor} even if we use PCA before applying IVA.   The results are very similar for TrojAI \Romannum{1} and TrojAI \Romannum{2} showing the robustness and generalizability of our method .

\begin{table}[h!]
\centering
\scriptsize 
\resizebox{\columnwidth}{!}{%
    \begin{tabular}{lrrrrrrrrr}\toprule
        &\multicolumn{3}{c}{\textbf{IVA}} & \multicolumn{3}{c}{\textbf{PARAFAC2}}
        \\\cmidrule(r){2-4}\cmidrule(r){5-7}
         & Precision & Recall & Accuracy & Precision & Recall & Accuracy \\\midrule
        \textbf{MNIST}    
                & 0.91 & 0.89 & 0.91
                & 0.93 & 0.91 & \textbf{0.92} \\

        \textbf{CIFAR-10}    
                & 0.86 & 0.84 & 0.85
                & 0.89 & 0.87 & \textbf{0.88} \\
                
        \textbf{TrojAI I-R50}  
                & 0.86 & 0.86 & 0.84
                & 0.90 & 0.86 & \textbf{0.87} \\
                
        \textbf{TrojAI II-R50}      
                & 0.91 & 0.78 & 0.82
                & 1.00 & 0.78 & \textbf{0.87} \\
        \textbf{TrojAI I-D121}      
                & 0.76 & 0.81 & 0.79
                & 0.82 & 0.88 & \textbf{0.85} \\
                
        \textbf{TrojAI II-D121}    
                & 0.73 & 0.80 & 0.79
                & 0.82 & 0.90 & \textbf{0.83}\\
        \textbf{TrojAI I-Iv3}     
                & 0.73 & 0.73 & 0.78
                & 0.75 & 0.82  & \textbf{0.81}\\
        \textbf{TrojAI II-Iv3}     
                & 0.77 & 0.77 & 0.80
                & 0.79 & 0.85  & \textbf{0.83}
                
        \\\bottomrule
    \end{tabular}
}
\caption{Performance metrics: Precision, Recall and Accuracy for 2 Algorithms: IVA, and PARAFAC2 using MNIST, CIFAR-10, TrojAI I, and TrojAI II datasets.}\label{Tab2}
\end{table}

\subsection{Clustering analysis}

From IVA and PARAFAC2, we get both the sources or components and the mixing matrix $\mathbf{A}$. Each column of $\mathbf{A}\in \mathbb{R}^{K\times 1}$ shows the contribution of each DNN model to one source. We have applied 2-means clustering using the first two columns of $\mathbf{A}$, meaning the contribution of $K$ models in the first two sources.
 Thus, the dimension of $\mathbf{A}$ before clustering is the same for same dataset, but it differs across the dataset based on the number of models in that dataset. As an example, we  plotted the clustering results using PARAFAC2 on TrojAI \Romannum{1} - R50 models in Figure 4. Here, red circles are cluster 1 consisting of trojan (backdoor) models while blue circles indicate clean models. As expected from our previous correlation analysis, all the backdoor models reside very close to each other in one cluster and the clean models form a sparse cluster. The blue circles in the red cluster depict false positives and red circles in the blue cluster are false negatives. To summarize our clustering results, we compute the mean silhouette scores \cite{shahapure2020cluster}; ranging from –1 to 1 where scores close to 1 indicate that a point is well correlated to its own cluster and poorly matched to other clusters. Table 2 shows that the silhouette score is greater than 0.65 in all scenarios, hence the DNN models well fitted in two clusters. However, PARAFAC2 has better scores, as we do not need to reconstruct the mixing matrix, $\mathbf{A}$, because we do not apply PCA first. But in IVA ,  $\mathbf{A}^{[k]}$ is back reconstructed using $\mathbf{A}^{[k]}={(\mathbf{D}^{[k]})^{-T}}$ ${\mathbf{\hat{A}}^{[k]}}$ where $\mathbf{D}^{[k]}$ is the data reduction matrix and ${\mathbf{\hat{A}}^{[k]}}$ is estimated mixing matrix from IVA decomposition, resulting in  some information loss.

\begin{figure}[h!]
  
  \centering
  \includegraphics[scale=.46]{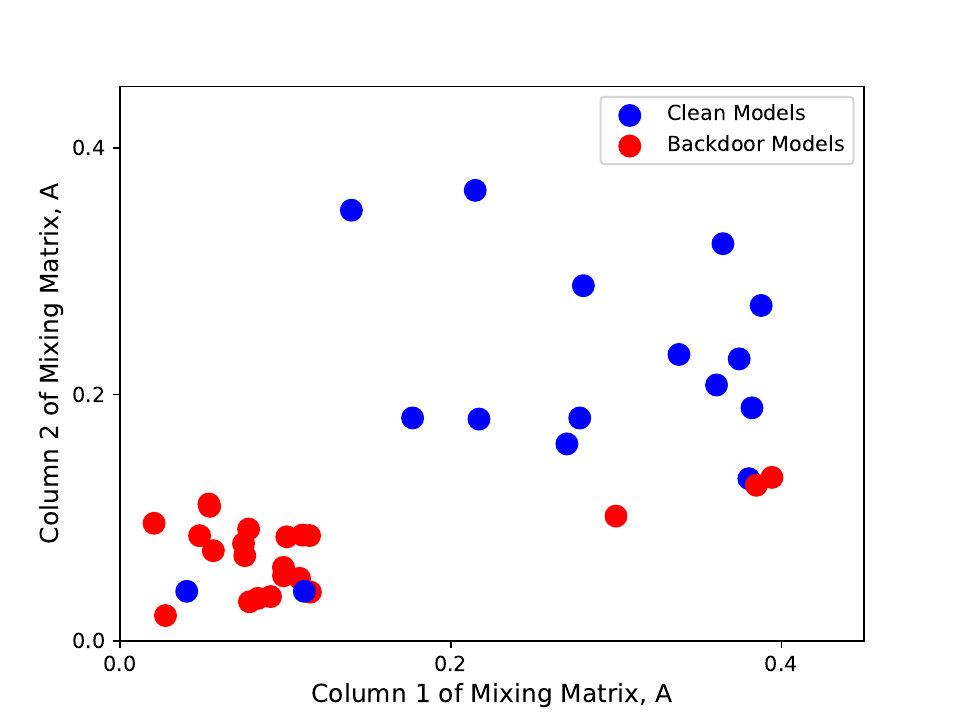}
  \caption{2-means clustering using PARAFAC2 on TrojAI I-R50 models. Red circles are trojan and blue circles are clean. }
\end{figure}

\begin{table}[h!]
\resizebox{\columnwidth}{0.7cm}{%

        \begin{tabular}{lrrrrrrrr}\toprule
               
             &MNIST&CIFAR-10&R50-\Romannum{1}&R50-\Romannum{2}&D121-\Romannum{1}&D121-\Romannum{2}&Iv3-\Romannum{1}&Iv3-\Romannum{2}\\\midrule

             \textbf{IVA}    & 0.81  &0.79& 0.75& 0.74  & 0.72& 0.73  & 0.71& 0.73     \\
                    
            \textbf{PARAFAC2}   & 0.84  &0.82& 0.81& 0.82  & 0.77& 0.76  & 0.75& 0.76      \\
              \hline

        \end{tabular}
        }
        \caption{Mean Silhouette scores for IVA, and PARAFAC2. Higher score means clusters are well apart from each other and clearly distinguished.}\label{Tab2}
    \end{table}

\subsection{Comparison with other  methods}

We compare the performance of our method with four SOTA backdoor detection methods: NC \cite{wang2019neural}, ABS \cite{liu2019abs}, Universal Litmus Patterns (ULP) \cite{kolouri2020universal},  Activation Clustering (AC) \cite{chen2018detecting}, and K-Arm \cite{shen2021backdoor}. For robustness measures of the pipelines, we also compute confidence intervals on the empirical accuracies using the standard equation for binomial proportions,  confidence  interval=$z\times \sqrt{(accuracy\times (1-accuracy))/n}$.  $n$ is the number of  models classified as backdoored or clean, and we use $z$ = 1.96 and thus have 95\% confidence intervals \cite{witten2002data}. We use the same batch size for the optimization-based methods, such as NC, ABS and ULP, and K-Arm to allow for fair comparison.

\noindent As illustrated in Table 3, our pipeline, using PARAFAC2, consistently outperforms competing methods across MNIST, CIFAR-10, and TrojAI datasets for various architecture types. Specifically, for MNIST CNNs, PARAFAC2 registers an accuracy of 92.07\% and an ROC-AUC of 0.93, an improvement of  
$\sim$1\% over the closest competitor, K-Arm. This superior performance is also evident in CIFAR-10 models.

\noindent For both TrojAI datasets, not only does PARAFAC2 surpass other methods in accuracy and ROC-AUC metrics, but our pipeline also exhibits narrower confidence intervals, suggesting robustness over alternative approaches. It's worth noting that performance metrics, universally across methods, are somewhat subdued on TrojAI datasets compared to MNIST and CIFAR-10 CNNs. This can be attributed to several factors: (i) TrojAI datasets exhibit greater trigger variations, encompassing aspects such as size, color, and location; (ii) the depth of TrojAI models is considerably more than our custom CNNs; and (iii) TrojAI employs multi-class poisoning, in contrast to the single-class poisoning leveraged for MNIST CNNs. Despite these intricacies, our methodology maintains an edge over all competitors.

\begin{table}[h!]
\fontsize{20}{18}\selectfont  
\setlength\tabcolsep{5pt}  
\renewcommand{\arraystretch}{2}  
\resizebox{\columnwidth}{!}{

        \begin{tabular}{rrrrrrrrrrrrr}\toprule
            & \multicolumn{2}{c}{{\textbf{NC}}}&\multicolumn{2}{c}{{\textbf{ABS}}}&\multicolumn{2}{c}{{\textbf{ULP}}}&\multicolumn{2}{c}{{\textbf{AC}}}&\multicolumn{2}{c}{{\textbf{K-Arm}}}&\multicolumn{2}{c}{{\textbf{PARAFAC2 (ours)}}}
            \\\cmidrule(r){2-3}\cmidrule(r){4-5}\cmidrule(r){6-7}\cmidrule(r){8-9}\cmidrule(r){10-11}\cmidrule(r){12-13}   
             &ROC-AUC&Acc&ROC-AUC&Acc&ROC-AUC&Acc&ROC-AUCC&Acc&ROC-AUC&Acc&ROC-AUC&Acc\\\midrule

             \textbf{MNIST}  & 0.83&0.84$\pm 0.12$ & 0.83&0.82$\pm 0.10$ &0.88&0.86$\pm 0.09$ & 0.65&0.66$\pm 0.15$ &0.92&0.92$\pm 0.06$ &\textbf{0.93}& \textbf{0.92$\pm \textbf{0.06}$} \\
             \textbf{CIFAR-10}  & 0.82&0.84$\pm 0.10$ & 0.83&0.83$\pm 0.11$ &0.86&0.85$\pm 0.07$ & 0.61&0.62$\pm 0.14$ &0.87&0.86$\pm 0.05$ &\textbf{0.88}& \textbf{0.88$\pm \textbf{0.04}$} \\

           \textbf{TrojAI I-R50}    &0.74&  0.73$\pm 0.14$ &0.78&0.76$\pm 0.14$ &0.82& 0.81$\pm 0.12$ &0.57& 0.57$\pm 0.15$ &0.88& 0.86$\pm 0.10$ &\textbf{0.89}& \textbf{0.87$\pm \textbf{0.09}$} \\
                    
            \textbf{TrojAI II-R50}    &0.74& 0.72$\pm 0.13$ &0.75& 0.75$\pm 0.13$  &0.81&0.80$\pm 0.11$ &0.60& 0.58$\pm 0.14$& 0.87& 0.87$\pm 0.10$ &\textbf{0.87}& \textbf{0.87$\pm \textbf{0.10}$} \\
                   
            \textbf{TrojAI I-D121}    &0.72& 0.72$\pm 0.15$  &0.74&0.73$\pm 0.14$ &0.79& 0.77$\pm 0.13$ &0.55&0.56$\pm 0.17$ &0.84&0.83$\pm 0.12$ &\textbf{0.86}& \textbf{0.85$\pm \textbf{0.11}$} \\

           \textbf{TrojAI II-D121}    &0.74& 0.73$\pm 0.16$ &0.73&0.74$\pm 0.15$ &0.78& 0.76$\pm 0.14$ &0.55& 0.54$\pm 0.18$ &0.84& 0.83$\pm 0.12$ &\textbf{0.85}& \textbf{0.83$\pm \textbf{0.11}$} \\

            \textbf{TrojAI I-Iv3}    &0.77& 0.75$\pm 0.16$ &0.74&0.73$\pm 0.14$ &0.80& 0.78$\pm 0.13$ &0.58& 0.58$\pm 0.18$  &0.81& 0.81$\pm 0.11$&\textbf{0.83}& \textbf{0.83$\pm \textbf{0.10}$}\\
            \textbf{TrojAI II-Iv3}    &0.75& 0.74$\pm 0.15$ &0.74&0.73$\pm 0.14$ &0.78& 0.78$\pm 0.13$ &0.55& 0.54$\pm 0.17$  &0.85& 0.83$\pm 0.11$ &\textbf{0.86}& \textbf{0.83$\pm \textbf{0.11}$}\\\bottomrule
        \end{tabular}
        }
        \caption{Comparison of backdoor detection performance using
six methods including PARAFAC2 (ours)  and ours give better
performance in all three datasets and architectures.}\label{Tab2}
    \end{table}

\subsection{Scalability of our method }

Because backdoor detection may become a routine element of ML
operations, it is important that such methods are efficient.  Table 4
shows the time  required to make decisions about all of the
networks in one dataset for NC, ABS, ULP,  AC, K-Arm, and the two variants of
our methods.  However, our method  tends to be faster than NC, ABS, ULP, and K-Arm by an order
of magnitude.  This is due to the fact that our approach is
model agnostic and only needs to final fully connected layer
activations to make accurate classification decisions.  The running
time of AC is comparable to ours, but as noted in Table 3 it
is significantly less accurate.  As a result, our method is able to strike a balance
between efficiency and accuracy that is not present in any of the
competing algorithms.

\begin{table}[h!]
    \centering
    \scriptsize 
    \begin{tabular}{l rrrr}\toprule
        &\multicolumn{4}{c}{\textbf{computation time (s)}}
        \\\cmidrule(r){2-5}
        \textbf{Method} & MNIST & CIFAR-10 & TrojAI I & TrojAI II \\\midrule
        NC & 1346 & 1745 & 2117 & 2256 \\
        ABS & 1565 & 1970 & 2234 & 2437 \\
        ULP & 2514 & 2918 & 3678 & 3944 \\
        AC & 267 & 415 & 342 & 398 \\
        K-Arm & 1463 & 1696 & 2157 & 2235 \\
        IVA & 161 & 179 & 198 & 213 \\
        PARAFAC2 & 178 & 218 & 241 & 230 \\
        \bottomrule
    \end{tabular}
    \caption{Computation time including our two algorithms: IVA, PARAFAC2 and NC, ABS, ULP, AC, and K-Arm.}\label{Tab2}
\end{table}

\section{Conclusion}

Due to the wide and growing adoption of very large deep models, there is more risk of attacks on models trained on third party hardware and open access models.  In this paper, we have proposed a novel pipeline which uses tensor decomposition techniques for detecting trojaned models.  We have done extensive experiments using MNIST, CIFAR-10 and TrojAI data and show that our method exhibits better performance than SOTA methods with higher efficiency.



\pagebreak

\bibliographystyle{IEEEtran}
\bibliography{strings,refs}

\begin{thebibliography}{10}
\providecommand{\url}[1]{#1}
\csname url@samestyle\endcsname
\providecommand{\newblock}{\relax}
\providecommand{\bibinfo}[2]{#2}
\providecommand{\BIBentrySTDinterwordspacing}{\spaceskip=0pt\relax}
\providecommand{\BIBentryALTinterwordstretchfactor}{4}
\providecommand{\BIBentryALTinterwordspacing}{\spaceskip=\fontdimen2\font plus
\BIBentryALTinterwordstretchfactor\fontdimen3\font minus
  \fontdimen4\font\relax}
\providecommand{\BIBforeignlanguage}[2]{{%
\expandafter\ifx\csname l@#1\endcsname\relax
\typeout{** WARNING: IEEEtran.bst: No hyphenation pattern has been}%
\typeout{** loaded for the language `#1'. Using the pattern for}%
\typeout{** the default language instead.}%
\else
\language=\csname l@#1\endcsname
\fi
#2}}
\providecommand{\BIBdecl}{\relax}
\BIBdecl

\bibitem{Goodfellow2015}
\BIBentryALTinterwordspacing
I.~Goodfellow, J.~Shlens, and C.~Szegedy, ``Explaining and harnessing
  adversarial examples,'' in \emph{International Conference on Learning
  Representations}, 2015. [Online]. Available:
  \url{http://arxiv.org/abs/1412.6572}
\BIBentrySTDinterwordspacing

\bibitem{wang2019neural}
B.~Wang, Y.~Yao, S.~Shan, H.~Li, B.~Viswanath, H.~Zheng, and B.~Y. Zhao,
  ``Neural cleanse: Identifying and mitigating backdoor attacks in neural
  networks,'' in \emph{IEEE Symposium on Security and Privacy}.\hskip 1em plus
  0.5em minus 0.4em\relax IEEE, 2019.

\bibitem{karra2020trojai}
K.~Karra, C.~Ashcraft, and N.~Fendley, ``The trojai software framework: An
  opensource tool for embedding trojans into deep learning models,''
  \emph{arXiv preprint arXiv:2003.07233}, 2020.

\bibitem{anderson2011joint}
M.~Anderson, T.~Adali, and X.-L. Li, ``Joint blind source separation with
  multivariate gaussian model: Algorithms and performance analysis,''
  \emph{IEEE Transactions on Signal Processing}, vol.~60, no.~4, 2011.

\bibitem{hossain2022data}
K.~M. Hossain, S.~Bhinge, Q.~Long, V.~D. Calhoun, and T.~Adali, ``Data-driven
  spatio-temporal dynamic brain connectivity analysis using falff: application
  to sensorimotor task data,'' in \emph{2022 56th Annual Conference on
  Information Sciences and Systems (CISS)}.\hskip 1em plus 0.5em minus
  0.4em\relax IEEE, 2022.

\bibitem{kiers1999parafac2}
H.~A. Kiers, J.~M. Ten~Berge, and R.~Bro, ``Parafac2—part i. a direct fitting
  algorithm for the parafac2 model,'' \emph{Journal of Chemometrics: A Journal
  of the Chemometrics Society}, vol.~13, no. 3-4, 1999.

\bibitem{acar2022tracing}
E.~Acar, M.~Roald, K.~M. Hossain, V.~D. Calhoun, and T.~Adali, ``Tracing
  evolving networks using tensor factorizations vs. ica-based approaches,''
  \emph{Frontiers in neuroscience}, vol.~16, 2022.

\bibitem{morcos2018insights}
A.~Morcos, M.~Raghu, and S.~Bengio, ``Insights on representational similarity
  in neural networks with canonical correlation,'' \emph{Advances in Neural
  Information Processing Systems}, vol.~31, 2018.

\bibitem{cortes2012algorithms}
C.~Cortes, M.~Mohri, and A.~Rostamizadeh, ``Algorithms for learning kernels
  based on centered alignment,'' \emph{The Journal of Machine Learning
  Research}, vol.~13, 2012.

\bibitem{raghu2017svcca}
M.~Raghu, J.~Gilmer, J.~Yosinski, and J.~Sohl-Dickstein, ``Svcca: Singular
  vector canonical correlation analysis for deep learning dynamics and
  interpretability,'' \emph{Advances in neural information processing systems},
  2017.

\bibitem{ailon2009fast}
N.~Ailon and B.~Chazelle, ``The fast johnson--lindenstrauss transform and
  approximate nearest neighbors,'' \emph{SIAM Journal on computing}, vol.~39,
  no.~1, 2009.

\bibitem{eftekhari2011two}
A.~Eftekhari, M.~Babaie-Zadeh, and H.~A. Moghaddam, ``Two-dimensional random
  projection,'' \emph{Signal processing}, vol.~91, no.~7, pp. 1589--1603, 2011.

\bibitem{adali2014diversity}
T.~Adali, M.~Anderson, and G.-S. Fu, ``Diversity in independent component and
  vector analyses: Identifiability, algorithms, and applications in medical
  imaging,'' \emph{IEEE Signal Processing Magazine}, vol.~31, no.~3, 2014.

\bibitem{bro1999parafac2}
R.~Bro, C.~A. Andersson, and H.~A. Kiers, ``Parafac2—part ii. modeling
  chromatographic data with retention time shifts,'' \emph{Journal of
  Chemometrics: A Journal of the Chemometrics Society}, vol.~13, no. 3-4, 1999.

\bibitem{taylor1990interpretation}
R.~Taylor, ``Interpretation of the correlation coefficient: a basic review,''
  \emph{Journal of diagnostic medical sonography}, vol.~6, no.~1, 1990.

\bibitem{weisstein2004bonferroni}
E.~W. Weisstein, ``Bonferroni correction,'' \emph{https://mathworld. wolfram.
  com/}, 2004.

\bibitem{chen2018detecting}
B.~Chen, W.~Carvalho, N.~Baracaldo, H.~Ludwig, B.~Edwards, T.~Lee, I.~Molloy,
  and B.~Srivastava, ``Detecting backdoor attacks on deep neural networks by
  activation clustering,'' \emph{arXiv preprint arXiv:1811.03728}, 2018.

\bibitem{sidiropoulos2017tensor}
N.~D. Sidiropoulos, L.~De~Lathauwer, X.~Fu, K.~Huang, E.~E. Papalexakis, and
  C.~Faloutsos, ``Tensor decomposition for signal processing and machine
  learning,'' \emph{IEEE Transactions on Signal Processing}, vol.~65, no.~13,
  pp. 3551--3582, 2017.

\bibitem{shahapure2020cluster}
K.~R. Shahapure and C.~Nicholas, ``Cluster quality analysis using silhouette
  score,'' in \emph{2020 IEEE 7th International Conference on Data Science and
  Advanced Analytics (DSAA)}.\hskip 1em plus 0.5em minus 0.4em\relax IEEE,
  2020, pp. 747--748.

\bibitem{liu2019abs}
Y.~Liu, W.-C. Lee, G.~Tao, S.~Ma, Y.~Aafer, and X.~Zhang, ``Abs: Scanning
  neural networks for back-doors by artificial brain stimulation,'' in
  \emph{Proceedings of the 2019 ACM SIGSAC Conference on Computer and
  Communications Security}, 2019, pp. 1265--1282.

\bibitem{kolouri2020universal}
S.~Kolouri, A.~Saha, H.~Pirsiavash, and H.~Hoffmann, ``Universal litmus
  patterns: Revealing backdoor attacks in cnns,'' in \emph{Proceedings of the
  IEEE/CVF Conference on Computer Vision and Pattern Recognition}, 2020.

\bibitem{shen2021backdoor}
G.~Shen, Y.~Liu, G.~Tao, S.~An, Q.~Xu, S.~Cheng, S.~Ma, and X.~Zhang,
  ``Backdoor scanning for deep neural networks through k-arm optimization,'' in
  \emph{International Conference on Machine Learning}.\hskip 1em plus 0.5em
  minus 0.4em\relax PMLR, 2021, pp. 9525--9536.

\bibitem{witten2002data}
I.~H. Witten and E.~Frank, ``Data mining: practical machine learning tools and
  techniques with java implementations,'' \emph{Acm Sigmod Record}, vol.~31,
  no.~1, 2002.

\end{thebibliography}

\end{document}